\documentclass[11pt, a4paper, copyright, nonumbering]{main}
\usepackage[numbers, sort&compress]{natbib}

\usepackage{dblfloatfix}
\usepackage{ulem}
\usepackage{caption}
\usepackage{dramatist}
\usepackage{xspace}
\usepackage{pifont}
\usepackage{multirow}
\usepackage{tcolorbox}
\usepackage{xltabular}
\usepackage{longtable}
\usepackage{pdfpages}
\definecolor{linkcolor}{rgb}{0.956,0.298,0.235}
\definecolor{citecolor}{HTML}{1976D2}
\usepackage{tabularx}
\usepackage{hyperref}
\usepackage{graphicx}
\usepackage{subfigure}
\usepackage{amsmath}
\usepackage{amssymb}
\usepackage{array}
\usepackage{booktabs}
\usepackage{multirow}
\usepackage{arydshln}
\usepackage{fontawesome}
\usepackage[utf8]{inputenc}
\hypersetup{
    colorlinks=true,
    linkcolor=citecolor,
    filecolor=magenta,
    urlcolor=cyan,
    citecolor=citecolor,
}

\usepackage{amsfonts}
\usepackage{amsmath}
\usepackage{amssymb}
\usepackage{lineno}

\usepackage[bottom]{footmisc}

\usepackage{CJKutf8}
\usepackage{subfigure}
\usepackage{setspace}

\usepackage{xspace}

\makeatletter
\def\@BTrule[#1]{
  \ifx\longtable\undefined
    \let\@BTswitch\@BTnormal
  \else\ifx\hline\LT@hline
    \nobreak
    \let\@BTswitch\@BLTrule
  \else
     \let\@BTswitch\@BTnormal
  \fi\fi
  \global\@thisrulewidth=#1\relax
  \ifnum\@thisruleclass=\tw@\vskip\@aboverulesep\else
  \ifnum\@lastruleclass=\z@\vskip\@aboverulesep\else
  \ifnum\@lastruleclass=\@ne\vskip\doublerulesep\fi\fi\fi
  \@BTswitch}
\makeatother

\addto\extrasenglish{

}

\newlength\savewidth\newcommand\shline{\noalign{\global\savewidth\arrayrulewidth
  \global\arrayrulewidth 1pt}\hline\noalign{\global\arrayrulewidth\savewidth}}

\reportnumber{001}
\renewcommand{\today}{}
\newcommand{\model}{UniViTAR}

\newcommand{\fontgray}{\textcolor{gray!40}}

\title{\centering 
UniViTAR: Unified Vision Transformer with Native Resolution
}

\author[*]{
\small Limeng Qiao~~~~Yiyang Gan~~~~Bairui Wang~~~~Jie Qin~~~~Shuang Xu~~~~Siqi Yang~~~~Lin Ma$^{\ddagger}$

\small Meituan

\small Project Page: \ \url{https://github.com/MM-MVR/UniViTAR}
\vspace{-6mm}
}

\begin{abstract}
Conventional Vision Transformer simplifies visual modeling by standardizing input resolutions, often disregarding the variability of natural visual data and compromising spatial-contextual fidelity. 
While preliminary explorations have superficially investigated native resolution modeling, existing approaches still lack systematic analysis from a visual representation perspective.
To bridge this gap, we introduce \model, a family of homogeneous vision foundation models tailored for unified visual modality and native resolution scenario in the era of multimodal.
Our framework first conducts architectural upgrades to the vanilla paradigm by integrating multiple advanced components. Building upon these improvements, a progressive training paradigm is introduced, which strategically combines two core mechanisms: (1) resolution curriculum learning, transitioning from fixed-resolution pretraining to native resolution tuning, thereby leveraging ViT’s inherent adaptability to variable-length sequences, and (2) visual modality adaptation via inter-batch image-video switching, which balances computational efficiency with enhanced temporal reasoning.
In parallel, a hybrid training framework further synergizes sigmoid-based contrastive loss with feature distillation from a frozen teacher model, thereby accelerating early-stage convergence.
Finally, trained exclusively on public datasets, externsive experiments across multiple model scales from 0.3B to 1B demonstrate its effectiveness.
\end{abstract}

\footnotetext{$^\ddagger$ \ Project Leader} 

\begin{document}
\begin{CJK*}{UTF8}{gbsn}

\maketitle

\vspace{4mm}
\begin{figure}[htb]
\centering
\includegraphics[width=\textwidth]{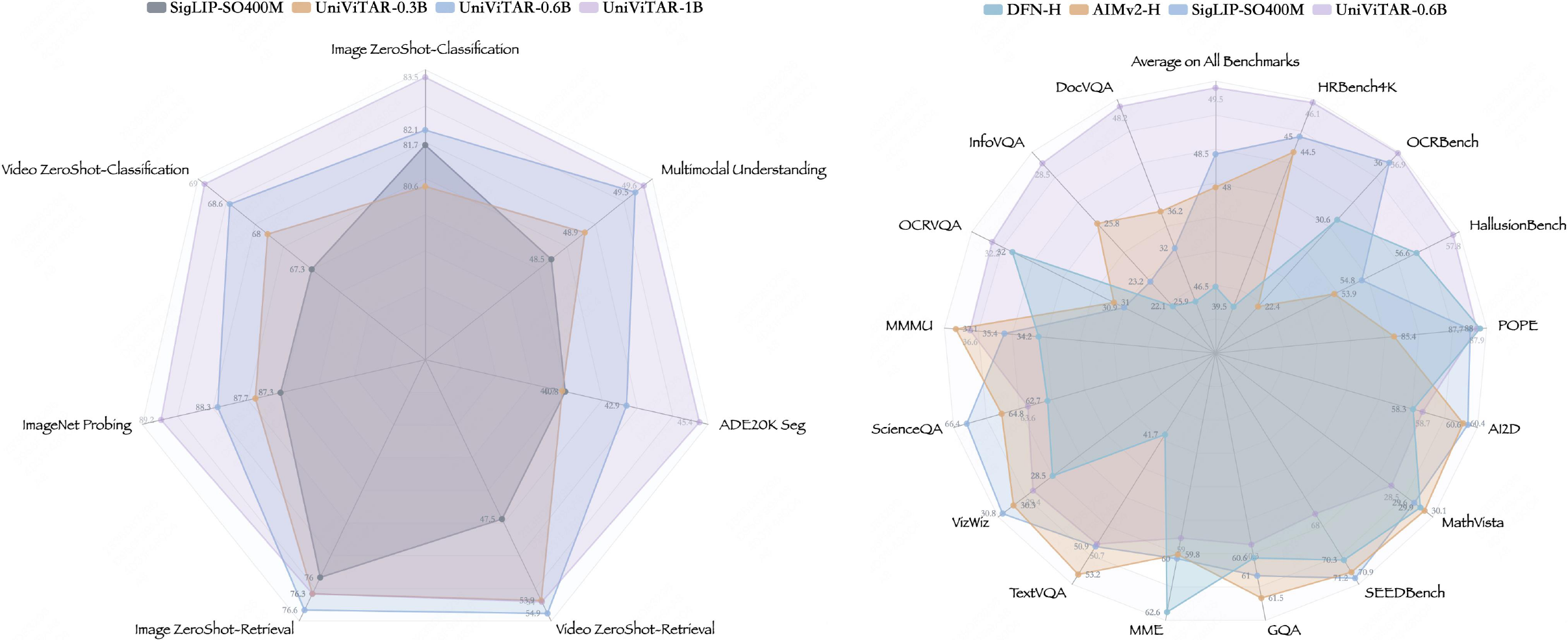}
\caption{
    An overview of the \model’s scaling behavior on a broad range of tasks (left) and its superior downstream multimodal performance (right) across diverse benchmarks.}
\label{fig:univitar_abstra}
\end{figure}

\section{Introduction}

In the era of rapid advancements of multimodal large models, Vision Transformer (ViT)~\cite{dosovitskiy2020image}, characterized by its simplicity and scalability, has emerged as a foundational architecture for visual representation learning. Drawing inspiration from transformer-based large language models, conventional ViT usually uniformly converts raw visual data into square aspect ratio and standardized resolution to reduce modeling complexity and simplify the training workflow. While this paradigm simplifies feature extraction and aligns with existing engineering practices, it inherently imposes artificial constraints on real-world visual data by disregarding the inherent variability of natural images. Real-world visual content exists across diverse resolutions and aspect ratios, with native shape often preserving finer spatial details and contextual information.

Recent studies have early explored the vision backbone under the native resolution paradigm. FlexViT~\cite{beyer2023flexivit} introduces a flexible ViT architecture featuring dynamical patch size selection in the patch embedding layer, which facilitates smooth variation of token sequence length through parametric scaling. In contrast, NaViT~\cite{dehghani2023patch} maintains fixed patch size while directly processing native resolution images with varying aspect ratios, where the token sequence length of different images changes dynamically. This approach demonstrates the feasibility and benefits of adopting natural language processing style packing strategies for vision foundational model. Qwen-VL's~\cite{wang2024qwen2vlenhancingvisionlanguagemodels,bai2025qwen25vltechnicalreport} vision encoder inherits NaViT's core configuration while specifically investigating native resolution impacts from a multimodal large model perspective. While the aforementioned approaches have attracted initial research attention, the field still lacks a comprehensive series of architecture-homologous vision backbones that can simultaneously support native- and fixed-resolution processing, achieve high-fidelity feature extraction for both images and videos.
		
To address this gap, we present the \textbf{Uni}fied \textbf{Vi}sion \textbf{T}ransformer with N\textbf{A}tive \textbf{R}esolution, termed as \textbf{\model}, a family of vision foundational backbones designed to uniformly process visual modalities (image or video) with native resolution and dynamic aspect ratio. Building upon insights from large language model recent practices and architectural innovations in visual transformers, our approach firstly conduct systematic architectural upgrades to the vanilla ViT paradigm by integrating multiple advanced components: 2D Rotary Position Embedding, SwiGLU activation function, RMSNorm layer, QK-Norm mechanism, and LayerScale module. These modifications collectively establish a more robust architectural foundation compared to conventional implementations.
Secondly, we develop a progressive training paradigm with two complementary adaptation strategies: \textit{1)} the progressive resolution adaptation strategy employs curriculum learning from fixed low-resolution (e.g., 224) pretraining to native-resolution fine-tuning. Notably, our experiments reveal that the advanced ViT architecture exhibit remarkable adaptability - models pretrained at fixed resolution can efficiently generalize to variable-length visual sequences through limited native resolution tuning. \textit{2)} the progressive visual modality adaptation strategy addresses computational challenges in video processing by deferring video data integration to the final training phase. We further demonstrate that alternating image-video training sequences (inter-batch modality switching) significantly outperforms mixed-batch (intra-batch modality mixing) in preserving image understanding capabilities while acquiring temporal reasoning skills.
Thirdly, we implement a hybrid training framework combining contrastive learning objectives with distillation techniques. Our primary optimization employs a sigmoid-based contrastive loss~\cite{zhai2023sigmoid} for unified image-video representation learning. To accelerate early-stage convergence, we further incorporate feature distillation from a frozen vision teacher model as an auxiliary training objective during initial phases, then gradually phasing out this regularization as the model matures.
Finally, through this comprehensive approach trained on public-accessible datasets, we successfully scale a family of vision backbones supporting native resolutions and both visual modalities,  with parameter counts ranging from 0.3B to 1B.  Extensive evaluations demonstrate the effectiveness of our proposed methods.
		
\noindent Specifically, the contributions of our \model~family are summarized as follows:
\begin{itemize}
    \setlength{\itemsep}{5pt}
    \item We introduce a family of homogeneous visual foundation models that support native resolution and unified feature extraction across visual modalities, offering the community a versatile framework for multimodal research.
    \item We develop an efficient and effective progressive training strategy that addresses the computational challenges of native resolution modeling while systematically enhancing the model’s image-caption alignment capability.
    \item We train our models with public-accessible datasets, achieve leading performance with limited resources, and observe a trend of performance increasing with parameter scaling.
\end{itemize}

\begin{figure}[t]
\centering
\includegraphics[width=\textwidth]{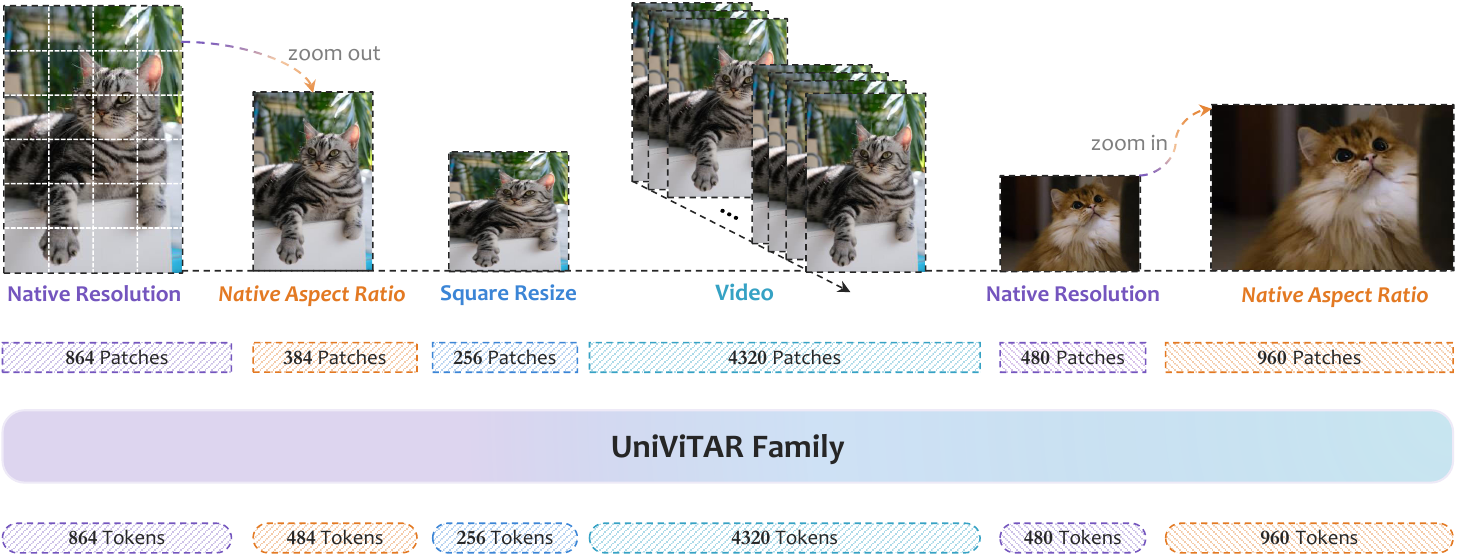}
\caption{
    \textbf{Brief illustration of \model~family pipeline}. \model~supports processing input at its native resolution, and also supports scaling the resolution down or up while maintaining the aspect ratio to accommodate different application scenarios, such as higher computational efficiency or finer-grained visual details. By treating video inputs as temporally extended image sequences, the framework uniformly produces longer variable-length visual token sequences.}
\label{fig:univitar_intra}
\end{figure}

\section{Related Work}

\subsection{Flexible Vision Transformers}
Vision Transformers have showcased impressive performance in numerous visual tasks, such as image classification \cite{dosovitskiy2020image}, image language pre-training \cite{radford2021clip}, etc. Those methods work only at a single, fixed resolution. Some works \cite{si2022inception, bai2023qwen} attempt to meet the need for fine-grained visual representation by adapting the model to a higher resolution during the fine-tuning stage. However, directly resizing the input to a fixed square resolution still limits their representation capacity in diverse visual scenarios. Recently, there are some works in vision transformers attempting to accommodate images with native resolutions with variable aspect ratios. ViTAR \cite{fan2024vitar} proposes an adaptive token merger module to alleviate the constraints of fixed resolution and adapt to multi-resolution inputs. However, it still limited by a predefied number of tokens that the model ultimately aims to obtain. NaViT \cite{dehghani2023patch} introduces sample packing used in language modeling for handling variable sequence length of image patches. Meanwhile, it introduces a factorized positional embedding schema in vanilla ViT to support variable aspect ratios and extrapolate to unseen resolutions. Qwen2.5-VL \cite{bai2025qwen25vltechnicalreport} integrates an NaViT-like approach to support native input resolutions, and employs multiple training phases for adapting it to multimodal large languages models, including CLIP pre-training, vision-language alignment, etc.

\subsection{Vision Foundation Models}
The development of vision foundation models has progressed through distinct phases, beginning with supervised learning paradigm dominated by landmark architectures like ResNet~\cite{he2016deep} and ViT~\cite{dosovitskiy2020image}, which established performance benchmarks through reliance on labeled data. However, the field witnessed a paradigm shift with the rise of self-supervised learning, which circumvented annotation bottlenecks through three principal branches: contrastive learning frameworks like SimCLR~\cite{chen2020simple} and MoCo~\cite{he2020momentum}, masked image modeling methods such as BEiT~\cite{wang2023image} and MAE~\cite{he2022masked}, and self-distillation techniques including BYOL~\cite{grill2020bootstrap} and DINO~\cite{caron2021emerging}. Recently, language-supervised contrastive pre-training has emerged as a transformative paradigm, exemplified by CLIP~\cite{radford2021learning}, which aligns multimodal embeddings through noise-robust contrastive objectives, enabling zero-shot task generalization. This approach has been further refined in works like SigLIP~\cite{zhai2023sigmoid}, which employs a more efficient sigmoid-base loss function while preserving cross-modal transfer capabilities. 
Besides images, a robust visual foundation model with effective video alignment capabilities serves as another critical building block. 
The existing strategies for training such models can be classified into three main paradigms: training on video-only data~\cite{rizve2024vidla,zhao2024videoprism,wang2022internvideo,xue2022clip}, utilizing multimodal data encompassing both video and image~\cite{wang2022internvideo,bai2023qwen,wang2024qwen2vlenhancingvisionlanguagemodels,bai2025qwen25vltechnicalreport}, and incorporating multimodal data that integrate video, images, audio and other modalities~\cite{wang2024internvideo2,srivastava2024omnivec}. VideoPrism~\cite{zhao2024videoprism} employs a two-stage video-only pretraining strategy: contrastive learning followed by token distillation, yet lacks image understanding. VidLA~\cite{rizve2024vidla} adapts CLIP~\cite{radford2021learning} via spatio-temporal attention on video-text data. InternVideo~\cite{wang2022internvideo} combines masked video modeling with alternating video/image-text pretraining, enhanced by cross-modal attention, while InternVideo2~\cite{wang2024internvideo2} extends this framework with audio/speech modalities for multimodal alignment. 

\subsection{Multimodal Large Language Models}
Recently, multimodal large language models (MLLMs) have witnessed significant advancements and rapid development~\cite{openai2024gpt4o,liu2023llava,li2024llava,bai2023qwen,bai2025qwen25vltechnicalreport,wang2024qwen2vlenhancingvisionlanguagemodels,chen2024internvl}. As a critical modality in MLLMs, visual input encounters inherent limitations when relying on conventional ViT with fixed resolutions, which may induce shape distortions, content blurring, and suboptimal handling of images/videos with diverse aspect ratios, high resolutions, or dynamic frame rates. To mitigate these challenges, the field has converged on two principal technical directions: 1) The tiling-based paradigm, as adopted by models like \cite{guo2024llava_uhd,li2024llava,chen2024internvl,wu2024deepseek}, decomposes ultra-high-resolution inputs into a varied number of fix-resolution tiles, and each tile is processed by a fixed-resolution vision encoder. As such, it enables MLLMs adaptivity to dynamic-resolution images without padding or shape-distorting resizing. However, the tile limits the model's ability to capture spatial information across different tiles and the primary subjects of the images are often fragmented, leading to the loss of spatial relationships and quantitative information. 2) native-resolution methodology, exemplified by models such as ~\cite{wang2024qwen2vlenhancingvisionlanguagemodels,liu2024points15buildingvisionlanguagemodel,deitke2024molmopixmoopenweights}, attempts to circumvent the limitations of the tiling-based paradigm by using native resolution input. However, they typically employ a pretrained fixed-resolution vision transformer (ViT) as the vision encoder, which leads to additional costs associated with adapting the ViT's distribution.  

\section{Method}

\subsection{\model: Homologous Visual Foundation Model}

\subsubsection{Architecture Design}

\model~is a Transformer-based encoder model that inherits the original architecture of the conventional Vision Transformer \cite{dosovitskiy2020image} but incorporates the following advanced modifications:

\begin{figure}[t]
	\centering
	\includegraphics[width=\textwidth]{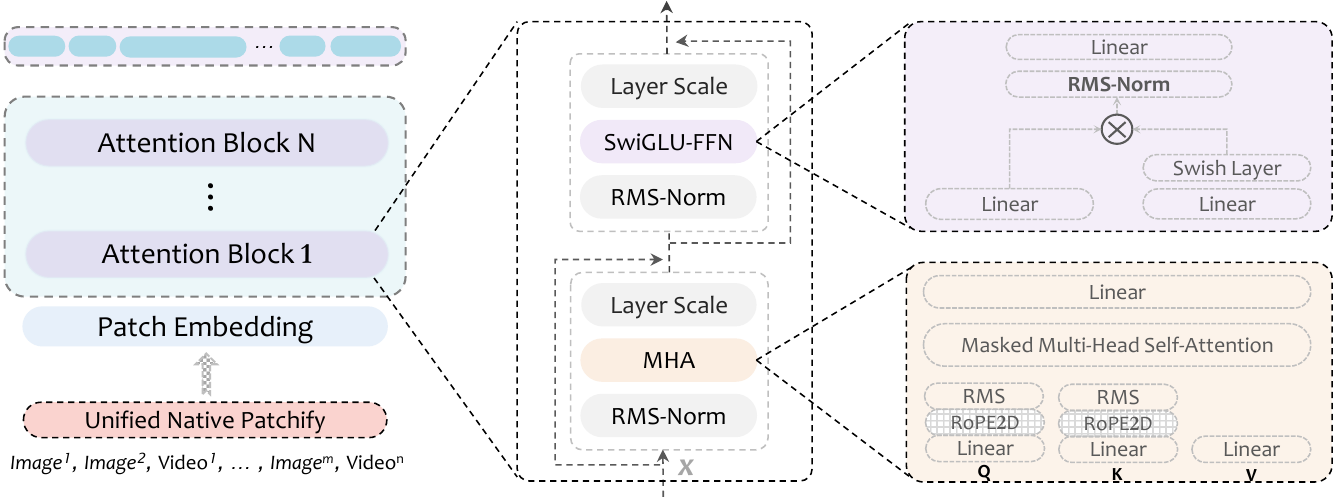}
	\caption{
		\textbf{Architecture of \model~family.} 
		All visual inputs are uniformly transformed into patch sequences and fed into Vision Transformer. In addition to using the \textit{Pre-Norm} approach, we also adopt \textit{RMS-Norm} as the normalization layer in both \textit{MHA} and \textit{FFN} module. Finally, a variable-length visual token sequence is produced.
		Best viewed on screen.    } 
	\label{fig:univitar_arch}
\end{figure}

\noindent \textbf{\textit{Unified Patchify for Native Image and Video Modality.}}
As illustrated in the Figure~\ref{fig:univitar_intra}, given the native input visual data $\mathbf{X} \in \mathbb{R}^{T \times C \times H \times W}$ of any vision modality (image, video), where $T=1$ represents image and $T > 1$ represents video, \textbf{\model}~ firstly patchifies $\mathbf{X}$ into a series of dynamic length visual patch sequences $\mathbf{P}=(N, S)$, where $N$ is the number of patches per image/video and $S$ is the number of pixels per patch.
Then a 3$D$ convolution layer is adopted as the \textit{Patch Embedding Layer} to consistently convert the above patch sequence into a visual token sequence $\mathbf{T}=(N, D)$, where $D$ is the hidden size of the following attention layers.

\noindent \textbf{\textit{2D Rotary Position Embedding.}}
Drawing on the architecture designs of language models, the original ViT regards the position information among different visual tokens as a one-dimensional association. In fact, considering that visual data usually has spatial association (row and column) and temporal association (time), the position information between different tokens is usually considered to be multi-dimensional.
Thence we remove the original absolute position encoding and introduce 2$D$-RoPE \cite{su2024roformer} into each subsequent encoder layer to capture the two-dimensional positional information of images.
Furthermore, we found that the presence or absence of the class token in the original ViT has almost no effect  on model performance. To ensure the consistency of position encoding, we also empirically remove the design of class token.

\noindent \textbf{\textit{SwiGLU and RMSNorm.}}
By leveraging the recent advances of LLaMA \cite{touvron2023llama} architecture design for language modeling, \model~incorporates SwiGLU as the feed-forward network (FFN) and replaces all normalization layers with RMSNorm. In addition, we adds an extra RMSNorm to each SwiGLU-FFN for good expressivity and improving the training stability.

\noindent \textbf{\textit{Query-Key Normalization.}}
In order to improve the stability of model training, we adopt the QK-Norm technique \cite{team2024chameleon, dehghani2023scaling} , which applies normalization to the queries and keys before the dot-product attention computation, to directly controls the norm growth of input to the softmax and avoid abnormal attention logits. Note that we still utilize RMSNorm as the norm function to ensure the consistency of the architecture.

\subsubsection{Homologous Model Scaling}
The \model~family consists of a comprehensive suite of foundational and scratch-train models, encompassing a parameter range from 0.3 to 1.4 billion, \textit{i.e.} \model-0.3B/0.6B/1B. The hyper-parameters and important information are listed in Table \ref{tab:univitar-mode-size} in details.

\begin{table}[t!]
	\caption{
		\textbf{Detailed architectural configuration for \model~family.}
	}
	\centering
	\renewcommand{\arraystretch}{1.2}
	\scalebox{0.85}{
		\begin{tabular}{l|c|c|c}
			\shline
			& \textbf{\model-0.3B} & \textbf{\model-0.6B} & \textbf{\model-1B} \\
			\shline
			Hidden Size & 1024 & 1280 & 1920 \\
			Intermediate Size & 4224 & 5184 & 7680 \\
			Num Layers & 24 & 32 & 32 \\
			Attention Heads & 16 & 16  & 24 \\
			Head Dimension & 64 & 80  & 80 \\
			\cdashline{1-4}
			\#Parameters (M) & 310 & 637  & 1419 \\
			\shline
		\end{tabular}
	}
	\label{tab:univitar-mode-size}
\end{table}

\subsection{Contrastive Vision-Language Pretrain with \model}

\subsubsection{Architecture Design}
In general, the acquisition of \model~largely follows the basic training paradigm of CLIP~\cite{radford2021clip}. Specifically, the native-resolution visual input $v$ is encoded into the visual feature space via the \model~encoder to obtain $F_v \in \mathbb{R}^{N_v \times D_v}$, while the textual input $t$ is projected into the textual feature space through a pretrained LLaMA  \cite{chen2024internvl} decoder to obtain $F_t \in \mathbb{R}^{N_t \times D_t}$. 
Note that the dynamic-length visual features $F_v $ are then uniformly converted into the visual embedding $f_v \in \mathbb{R}^{D_v}$ through a global average pooling layer and the feature corresponding to the \textit{<EOS>} token in $F_t$ is utilized as the textual representation $f_t \in \mathbb{R}^{D_t}$ of the input caption.
Subsequently, $f_v$ and $f_t$ are further projected into the same shared semantic feature space via a projection layer (typically a linear or cross-attention layer) respectively. 
Finally, through a contrastive loss function, the model semantically aligns the visual and textual features into the same feature space.
Instead of the communation-intensive softmax-based contrastive loss, we utilizes a simple pairwise sigmoid loss as the contrastive supervision objective~\cite{zhai2023sigmoid}, which operates solely on image-text pairs and does not require a global view of the pairwise similarities for normalization, conceptually decoupling the batch size from the definition of the contrastive task.
This semantic alignment process is usually ensured by leveraging a sufficiently large-scale dataset of image-text pairs  (typically on the magnitude of billions) to guarantee the effectiveness of the alignment.

\subsubsection{Optimized Contrastive Training Strategy}

Training CLIP-like model usually leads to significant computational resource requirements and even instability training problem. Especially under unified native-resolution settings, this dilemma will undoubtedly be further exacerbated. To alleviate this problem, we carefully design the training pipeline of \model~into four stages in sequence, as shown in Table \ref{tab:train-strategy}, to ensure that the model can converge efficiently and the training cost is controllable.

\begin{table}[t!]
\caption{
    \textbf{Detailed training strategy illustration of \model~family.}
    The proposed pipeline consists of four stages sequentially to ensure efficient model convergence and low training cost. Specifically, we apply distillation technology in Stage 1 to accelerate the model convergence process. Besides, in order to reduce the training cost, the text branch is only trained in Stage 3 with 1B samples, and the rest of the training stages are frozen. For \model~models of all sizes, a total of 14.6B data is trained in the entire training stage.
}
\centering
\renewcommand{\arraystretch}{1.2}
\scalebox{0.75}{
    
    \begin{tabular}{l|c|c|c|c}
        \shline
        & \textbf{Stage 1} & \textbf{Stage 2} & \textbf{Stage 3} & \textbf{Stage 4} \\
        \shline
        \multirow{5}{*}{Train Strategy} & 
          \begin{minipage}[t]{0.26\columnwidth}
              \vspace{-2mm} 
              \includegraphics[scale=0.5]{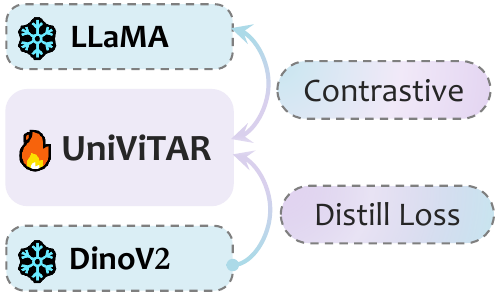} \label{fig: train-pipeline-s1}
              \vspace{-2.6mm} 
          \end{minipage} & 
          
          \begin{minipage}[t]{0.26\columnwidth}
            \vspace{2.2mm} 
            \includegraphics[scale=0.5]{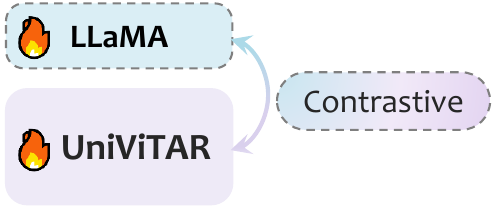} \label{fig: train-pipeline-s3}
          \end{minipage} & 

            \begin{minipage}[t]{0.26\columnwidth}
            \vspace{2.2mm} 
            \includegraphics[scale=0.5]{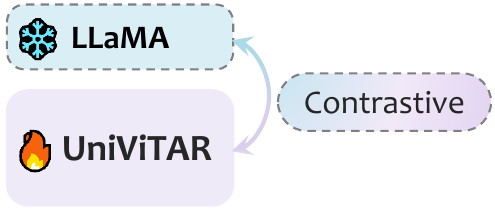} \label{fig: train-pipeline-s4}
          \end{minipage} &
          
          \begin{minipage}[t]{0.26\columnwidth}
            \vspace{2.2mm} 
            \includegraphics[scale=0.5]{figures/UniViTAR_CLIP_S4} \label{fig: train-pipeline-s4v2}
          \end{minipage} \\
          
          \cdashline{1-5}
          Data Modality & Image & Image& Image & Image, \textbf{Video} \\
          Resolution & $224 \times 224$ & $224 \times 224$ & \textbf{Native} & \textbf{Native} \\
            Vision Branch & Trainable & Trainable & Trainable & Trainable \\
            Text Branch & Frozen & Trainable & Frozen & Frozen \\
            Distill Branch & Frozen & - & - & - \\
          Loss Function & Sigmoid, KL & Sigmoid & Sigmoid & Sigmoid \\
          \cdashline{1-5}
          Seen Samples & 12B & 1B& 1B & 0.6B \\
        \shline
    \end{tabular}

}
\label{tab:train-strategy}
\end{table}

\noindent \textbf{\textit{Stage 1: Visual knowledge pre-acquisition with hybrid paradigm training.}} 
The primary objective of this phase is to efficiently pretrain a visual foundation model from scratch by integrating two classic learning paradigms:  vision-text contrastive learning and visual knowledge distillation.
Specifically, the model architecture adopts a \textbf{\textit{three-branch}} parallel structure comprising: 1) a scratch-trained  \model~model, 2) a well-pre-trained text branch, and 3) a well-pre-trained visual teacher branch. 
During this stage of training, only the target visual foundation model undergoes parameter updates, while the other two branches remain frozen throughout the process. This design ensures effective fusion of knowledge priors while reducing training costs as much as possible. Notably, our designed architecture employs LLaMA~\cite{chen2024internvl} as the text encoder and DINOv2-g~\cite{oquab2023dinov2}  as the visual teacher model, though these selections are not exclusive – any well-pre-trained foundation models could serve as viable alternatives. Regarding training objectives, we utilize a hybrid training paradigm of image-text contrast training guided by pre-trained visual knowledge as follows, 
\begin{equation}
\mathcal{L}_{overall} = \mathcal{L}_{contrastive}(f_v^{UniViTAR}, f_t^{LLaMA}) + \lambda \cdot \mathcal{L}_{distillation}(f_v^{UniViTAR}, f_v^{Dino})
\label{eq:stage-1-loss}
\end{equation}
where $\mathcal{L}_{contrastive}$ is the sigmoid loss from SigLIP~\cite{zhai2023sigmoid} and $\mathcal{L}_{distillation}$ is the KL Divergence~\cite{cover1999elements}.
The target visual foundation model functions as a visual knowledge bridge, simultaneously performing image-text alignment and image distillation. 
Although the training objective corresponding to Eq.~\ref{eq:stage-1-loss} enables scratched \model~to rapidly acquire visual knowledge and achieve well image-text alignment, the inherent inconsistency in optimization objective ultimately constrains the upper bound of alignment performance. Specifically, the distillation paradigm prioritizes preserving features from the visual teacher model, while the contrastive paradigm emphasizes building image-text aligned representations – this dual objective is sub-optimal for achieving optimal semantic alignment between images and texts. To address this, we strategically decay $\lambda$ in Eq.~\ref{eq:stage-1-loss} to 0 during later training phases, prioritizing the image-text alignment objective.

To accelerate convergence and enhance training stability, this stage exclusively employs single visual modality pre-training with all images resized to 224×224 resolution, further reducing computational costs. This phase processes 12B samples, constituting 82.2\% of the total training data (12B/14.6B). Notably, the $\lambda$ parameter is abruptly decayed to 0 at the 8B checkpoint.

\noindent \textbf{\textit{Stage 2: Finetune with full-parameter for superior alignment.}}
The objective of this stage is to further enhance the upper limit of image-text alignment and achieve a superior semantic-visual space by performing full parameter fine-tuning on both the vision backbone, \textit{i.e.} \model, and text branch. Notably, the visual distillation branch has been entirely removed in this phase. This stage utilizes exactly the same image-caption pair data as Stage 1, and all input shapes are resized to 224$\times$224. Considering the high computational cost of full parameter fine-tuning, the training process is conducted on 1B samples, accounting for 6.9\% of the total training data.

\noindent \textbf{\textit{Stage 3: Unlock the model-capacity of native-resolution.}}
In this stage, our strategy extends the model capability to handle native-resolution, thereby achieving robust image-text alignment for dynamic-resolution inputs. 
Architecturally, the core modules of vision transformer such as patch embedding layer, attention mechanism, and feed forward network inherently support both fixed- and native- resolution types. However, enabling native-resolution capacity of \model~necessitates addressing two critical challenges: (1) ensuring positional encodings are flexibly adaptable and thoroughly trained across variable sequence lengths, and (2) transfering feature distribution  from resized patches (from fixed-resolution training) to native patches (from original-resolution training) through efficient training. 
In practice, visual data is batched in its native form to preserve original resolutions and aspect ratios. Then the intra-batch images/videos are dynamically scaled (with aspect ratios maintained) to align total sequence lengths $L_{total}$ with a predefined token limit $L_{max}$. That is to say, when the value of $L_{max} / L_{total}$ is greater than 1, all data will be uniformly enlarged, and vice versa when it is less than 1, the shape of all data will be reduced, ensuring consistent computational loads across batches. Following visual transform and patchification, all patches from diverse images are concatenated into a single sequence, accompanied by a masking mechanism to delineate token boundaries for attention computation. Within attention blocks, each token’s receptive field is confined to tokens from the same image via masking, enabling isolated intra-image contextual modeling while preserving inter-sample independence. 
During training, resolution diversity within batches ensures comprehensive training of positional encodings across varying context lengths, progressively refining the model’s ability to generate features aligned with native patch distributions. At inference, inputs are processed directly at their native resolutions without resizing, leveraging the model’s trained adaptability to arbitrary resolutions for seamless image-text alignment. 
In this stage, 1B samples (6.85\% of the total training) were trained with the text branch frozen throughout the process. 

\noindent \textbf{\textit{Stage 4: Unifying visual modalities with image-video alternation training.}}
In this stage, our goal is to achieve the \model~that unifies image and video input modalities with native-resolution and dynamic video length.
Inspired by the InternVideo series~\cite{wang2022internvideo,wang2024internvideo2}, we utilize both video-text pairs and image-text pairs to optimize the \model~checkpoint derived from only image-text pairs, using an alternating training strategy of images and videos.
The rationale behind this strategy is threefold:
1) Although diversity and scale of video data are significantly lower than those of image data, the images contain general visual information as videos, which is good for learning visual content information in video data;
2) Incorporating image data helps avoid losing image understanding capabilities in the process of optimizing video comprehension capabilities;
3) The alternating training strategy ensures that parameters are updated with data from one modality at a time, allowing the model to focus on each modality individually.
For the alternating training strategy, we first shuffle the order of image and video data at the start of each epoch to increase randomness. The data is then divided according to the global batch size, with image and video batches interleaved in units of the global batch size. This ensures that each global batch contains only image data or video data during training.
To support native resolution and dynamic video length, given the maximum sample number of frame for each video $F_{max}$, we try to sample all frames of a video, which is denoted as $F$. When the frame number exceeds $F_{max}$, we uniformly sample $F_{max}$ frames from this video.
To unify the processing of images and videos, we duplicate image to simulate a two-frame video. 
Given the token limit $L_{max}$ and $L_{min}$ for each video clip, the pixel range for each frame can be calculated with $F$ and model patch size.
Next, each frame in the video is scaled to fit within the pixel range, while trying to maintain the original aspect ratio.
In this stage, a total of 695M data samples, comprising 630M images and 65M videos, were trained with the text branch frozen.

\subsection{\model~as a Vision Encoder for MLLMs.}
Vision encoders have emerged as pivotal components in modern multimodal large language models (MLLMs)~\cite{wang2024qwen2vlenhancingvisionlanguagemodels, mckinzie2024mm1, chen2024internvl}. 
In this section, we introduce a simple strategy for constructing an effective native resolution MLLM based on the \model~series. 
The common and industry-validated Vision-Language Models (VLMs) paradigm typically combines pretrained visual backbones with large language models, followed by multimodal training on a rich mixture of vision language tasks. 
To ensure fair comparison and minimize bias, we adhere to this established configuration. Specifically, we employ \model~as the vision encoder and employ Qwen2.5-1.5B as an efficient large language model (LLM). 
Following established practices~\cite{liu2023llava}, we implement a three-layer multilayer perceptron (MLP) with pre-normalization as the vision-language projector to bridge the visual and linguistic modalities. Furthermore, to address the quadratic computational complexity from high-resolution inputs, we apply a $2\times$ pixel-unshuffle operation~\cite{Shi_2016_CVPR} along the width dimension prior to projection.
		
During both training stage and inference stage, each input of image is resized to $28\times28$ patch multiples prior to ViT encoding, generating a dynamic-length vision token sequence.
For native-resolution MLLMs modeling, we identify two primary challenges. On one hand, due to the varying lengths of input samples, the boundary between vision and language tokens is not fixed. To enhance "modality isolation", we introduce specialized prompts, known as \textit{Boundary Markers}, such as {<image\_{start}>} and {<image\_{end}>}, at the beginning and end of the vision token sequence.  On the other hand, 2D-to-1D flattening of vision tokens may compromise the information of the height-width ratio. To mitigate this, we incorporate \textit{Line Anchors}, such as <line-\textit{idx}>, into the vision tokens, where \textit{idx} denotes the corresponding vertical positions in the original patchified image, thereby potentially strengthening positional awareness in compressed tokens. 
For a vision token sequence of length $h\cdot w$, the original arrangement $x^{<1, 1>}, ... ,x^{<1, w>}, ..., x^{<2, w>}, ..., x^{<h, w>}$ is transformed via \textit{Boundary Markers} and \textit{Line Anchors} into:
\begin{equation}
    \dashuline{\mathrm{<image\_start>}}, x^{<1,1>}, \dots ,x^{<1,w>}, \dashuline{\mathrm{<line-}1\mathrm{>}}, x^{<2,1>}, \dots, x^{<h,w>}, \dashuline{\mathrm{<line-}h\mathrm{>}}, \dashuline{\mathrm{<image\_end>}}
    \label{eq:arrangement}
\end{equation}
Notably, these added markers are string-based identifiers rather than special tokens within the tokenizer.
To systematically evaluate multimodal comprehension capabilities, we adopt a dual-stage training paradigm motivated by established methodologies in vision-language alignment~\cite{chen2024allava,chen2023sharegpt4v}. 
In the pretraining phase, we exclusively train the vision-language projector using large-scale image-caption pairs while freezing both the visual encoder and LLM parameters, thereby preserving their pretrained knowledge while optimizing cross-modal mapping. In the fine-tuning phase, we unfreeze the entire VLM and leverage diverse instruction-aware multimodal datasets to enable task-adaptive reasoning.

\section{Experiments}

This section first outlines the details related to training, beginning with implementation specifics such as data composition and hyperparameter configurations. 
Then we assess the model's effectiveness through comprehensive zero-shot evaluations on multiple image/video benchmarks. Subsequent analysis explores downstream potential across three key scenarios: linear-probing classification, pixel-level segmentation, and vision-language multimodal performance. Finally, ablation studies quantify the impact of critical architectural design choices in our methodology.

\subsection{Training Recipe}
\subsubsection{Data Details}

We collect public accessible image-text pairs and build our Merged-1B dataset. Table \ref{tab:univitar-image-data} summarizes the detailed composition of Merged-1B. Among them, DataComp-1B \cite{gadre2023datacomp}, COYO \cite{byeon2022coyo}, LAION-2B \cite{schuhmann2022laion5b}, LAION-400M \cite{schuhmann2021laion} are web-scale image-text pairs, DFN-2B \cite{byeon2022coyo} is with higher data quality released by \cite{fang2023data}, CC12M \cite{changpinyo2021cc12m} and CC3M \cite{sharma2018cc3m} consists of images with academic captions. Moreover, to further enhance the video feature extraction capabilities of \model, we meticulously constructed a dataset Merged-65M of roughly 65 million samples by randomly selecting video clips from three public accessible video datasets, \textit{i.e.}, Panda-70M~\cite{chen2024panda}, WebVid-10M~\cite{Bain21}, and InternVid-10M-FLT~\cite{wang2023internvid}. During training in the Stage 4, we mix about 630M image-text pairs in the video data and alternate images and videos for each batch iteration. Specifically, before each training epoch, we ensure that the data acquired within each global batch are all images or all videos, allowing the model to focus on each modality. We refer to the combined image and video data mentioned above as Merged-1.1B.

\begin{table}[htbp]
\caption{
    \textbf{Details of the training data for \model}. Note that Merged-1B and Merged-65M correspond to image and video modality respectively.
}
\centering
\tabcolsep=0.25cm
\renewcommand{\arraystretch}{1.2}
\scalebox{0.8}{
    \begin{tabular}{l|l|c|c|c|c|c}
        \shline
        Dataset & Source & Language & Samples & Total & Percentage & Used by  \\
        \shline
        \multirow{7}{*}{Merged-1B}
        & DataComp-1B \cite{gadre2023datacomp} & En & 408M & \multirow{7}{*}{1.08B} & 37.7\% & \multirow{7}{*}{Stage 1$\sim$4}\\
        & COYO \cite{byeon2022coyo} & En & 248M & & 22.9\% & \\
        & LAION-2B \cite{schuhmann2022laion5b} & En & 213M & & 19.7\% & \\
        & DFN-2B \cite{fang2023data} & En & 154M  & & 14.3\% & \\
        & LAION-400M \cite{schuhmann2021laion} & En & 52.7M  & & 4.9\% & \\
        & CC12M \cite{changpinyo2021cc12m} & En & 2.94M  & & 0.3\% & \\
        & CC3M \cite{sharma2018cc3m} & En & 2.32M & & 0.2\% & \\
        \cdashline{1-7}
        \multirow{3}{*}{Merged-65M}
        & Panda-70M~\cite{chen2024panda} & En & 52.1M & \multirow{3}{*}{65M}  & 80.2\% & \multirow{3}{*}{Stage 4} \\
        & WebVid-10M~\cite{Bain21} & En & 6.53M & & 10.0\% & \\
        & InternVid-10M-FLT~\cite{wang2023internvid} & En & 6.31M & & 9.8\% & \\
        \shline
    \end{tabular}
}
\label{tab:univitar-image-data}
\end{table}

\subsubsection{Hyperparameter Details}

The detailed hyperparameter configurations for each training stage are presented in Table \ref{tab:univitar-hyperparameter}. As tabulated, we utilize a progressive reduction of the peak learning rate in correlation with increasing visual backbone scale to ensure optimal training stability. Notably, the text modality learning rate in Stage 2 remains consistently one-tenth of the visual component throughout this phase. To enhance computational efficiency, we integrated the DeepSpeed optimization library~\cite{rasley2020deepspeed} employing three key strategies: ZeRO stage-1 optimizer sharding~\cite{rajbhandari2020zero}, gradient checkpointing~\cite{chen2016training}, and flash attention mechanisms~\cite{dao2022flashattention}. Additionally, the entire training process was conducted using the \textit{bfloat16} to maintain numerical stability.

\begin{table}[htb]
\caption{
    \textbf{Detailed training hyperparameter of \model~family}. Note that the symbol of $\to$ represents the peak learning rate and the minimum learning rate in the LR schedule.
}
\centering
\tabcolsep=0.3cm
\renewcommand{\arraystretch}{1.2}
\scalebox{0.80}{
    \begin{tabular}{r|c|c|c|c}
        \shline
        & \textbf{Stage 1} & \textbf{Stage 2} & \textbf{Stage 3} & \textbf{Stage 4} \\
        \shline
        Vision Encoder \textit{Init.} & Xavier \textit{init.}~\cite{glorot2010understanding}  & from Stage-1 & from Stage-2 & from Stage-3 \\
        Text Encoder \textit{Init.} & LLama~\cite{chen2024internvl} & LLama~\cite{chen2024internvl} & from Stage-2 & from Stage-3 \\
        Input Resolution & $224 \times 224$ & $224 \times 224$ & \textbf{Native} & \textbf{Native} \\
        Token Range & 256& 256 & 64 $\sim$ 16K & 64 $\sim$ 16K \\
        Global Batch Size & 32768 & 32768 & 32768 & $\sim$26K(Image), $\sim$4K(Video) \\
        Patch Dropout & 0.5 & 0.0 & 0.5 & 0.5 \\
        Warmup Steps & 2000& 2000& 2000& 1000\\
        Optimizer & AdamW & AdamW& AdamW& AdamW \\
        LR Schedule &  Cosine Decay & Cosine Decay & Cosine Decay & Cosine Decay \\
        \textit{0.3B} &  $1e^{-3} \to 1e^{-6}$ & $1e^{-5} \to 0$ & $1e^{-5} \to 0$ & $4e^{-6} \to 0$ \\
        \textit{0.6B} &  $1e^{-3} \to 1e^{-6}$ & $1e^{-5} \to 0$ & $1e^{-5} \to 0$ & $4e^{-6} \to 0$ \\
        \textit{1B} &  $8e^{-4} \to 1e^{-7}$ & $6e^{-6} \to 0$ & $6e^{-6} \to 0$ & $2e^{-6} \to 0$ \\
        \cdashline{1-5}
        Train Dataset & Merged-1B & Merged-1B & Merged-1B & Merged-1B, Merged-65M \\
        Seen Samples & 12B & 1B& 1B & 0.6B \\
        \shline
    \end{tabular}
    
}
\label{tab:univitar-hyperparameter}
\end{table}

\subsection{Results on Zero-shot Image Classification \& Retrieval}

\subsubsection{Evaluation Setup}
Our evaluation protocol encompasses both zero-shot classification and cross-modal retrieval tasks. 
For zero-shot classification, we conduct evaluation on ImageNet~\cite{deng2009imagenet} and its established variants~\cite{hendrycks2021natural,hendrycks2021many,recht2019imagenetv2,wang2019learning,barbu2019objectnet}. Each class is represented by multiple text prompts curated from~\cite{radford2021clip,zhai2022lit}. Class text embeddings are derived through averaging the embeddings generated by the text encoder across all class-specific prompts. We classify each image as the class that has the largest similarity with the image embedding. The \textit{Top-1} accuracy is utilized to evaluate the model performance.
For cross-modal retrieval assessment, we adopt the benchmark protocols defined in~\cite{karpathy2015deep}, evaluating on Flickr~\cite{young2014image} and MS-COCO ~\cite{lin2014microsoft} using their official partitions. The retrieval paradigm involves bidirectional image-text matching, namely image-to-text retrieval and text-to-image retrieval tasks. Note the performance is measured through \textit{Recall@1} metric.

\begin{table}[htb]
\centering
\setlength{\tabcolsep}{2pt}
\renewcommand{\arraystretch}{1.2}
\scriptsize
\caption{
    \textbf{Evaluation of zero-shot performance on various image benchmarks}. The symbol \faEyeSlash~indicates that the image-caption data used by the corresponding method is not publicly available. The blue shading corresponds to the zero-shot classification performance on ImageNet-variants, while the green is the average performance of zero-shot retrieval tasks.
}
\scalebox{0.96}{
    \begin{tabular}{rlccccccccccccc}
        \toprule
        \multirow{2}{*}{\textbf{Method}} & \multirow{2}{*}{\textbf{Data Source}} & \multirow{2}{*}{\textbf{Res.}} & \multirow{2}{*}{\textbf{Overall}} & \multicolumn{6}{c}{\textbf{ImageNet Variants}} & \multirow{2}{*}{\textbf{Overall}}& \multicolumn{2}{c}{\textbf{Flickr}} & \multicolumn{2}{c}{\textbf{COCO}} \\
        \cdashline{5-10} \cdashline{12-13} \cdashline{14-15}
        & & & & \textbf{IN-1K} & \textbf{IN-A} & \textbf{IN-R} & \textbf{IN-V2} & \textbf{IN-S} & \textbf{O-Net} & & \textbf{T$\to$I} & \textbf{I$\to$T} & \textbf{T$\to$I} & \textbf{I$\to$T} \\

        \midrule

        CLIP-L~\cite{radford2021clip} & WIT400M~\faEyeSlash & 224 &
        \cellcolor{blue!8}{72.1} & 75.5 & 70.8 & 87.8 & 69.8 & 59.6 & 68.9 & \cellcolor{green!8}{60.8} & 65.0 & 85.2 & 36.5 & 56.3\\

        SigLIP-L~\cite{zhai2023sigmoid} & WebLI10B-En~\faEyeSlash & 256 &
        \cellcolor{blue!8}{74.6} & 80.5 & 62.1 & 94.0 & 73.8 & 72.1 & 65.3 & \cellcolor{green!8}{73.5} & 79.0 & 91.8 & 52.3 & 70.8 \\

        OpenCLIP-L~\cite{openclip} & DataComp1B~ & 224 &
        \cellcolor{blue!8}{75.7} & 79.2 & 69.6 & 90.8 & 72.1 & 68.0 & 74.3 & \cellcolor{green!8}{67.9} & 73.4 & 89.0 & 45.7 & 63.3\\

        MetaCLIP-L~\cite{xu2023demystifying} & CC-2.5B & 224 &
        \cellcolor{blue!8}{76.6} & 79.2 & 72.3 & 92.1 & 72.6 & 69.0 & 74.6 & \cellcolor{green!8}{69.5} & 76.4 & 90.1 & 47.1 & 64.4 \\
        
        DFN-L~\cite{fang2023data} & DFN5B~\faEyeSlash & 224 &
        \cellcolor{blue!8}{77.1} & \textbf{82.2} & 67.5 & 91.8 & 75.7 & 70.4 & 74.8 & 
        \cellcolor{green!8}{69.8} & 75.5 & 89.6 & 48.6 & 65.6 \\

        EVA02-L~\cite{sun2023evaclip} & Merged-2B~ & 336 &
        \cellcolor{blue!8}{77.5} & 79.8 & 76.2 & 92.7 & 73.0 & 68.1 & 74.9 & 
        \cellcolor{green!8}{69.9} & 78.0 & 89.6 & 47.9 & 64.2 \\

        CLIPAv2-L~\cite{li2023clipa} & DataComp1B~ & 336 &
        \cellcolor{blue!8}{78.1} & 80.3 & 77.7 & 93.3 & 73.5 & 70.9 & 73.1 & 
        \cellcolor{green!8}{69.5} & 74.6 & 90.4 & 47.2 & 65.6 \\

        SigLIP-L~\cite{zhai2023sigmoid} & WebLI10B-En~\faEyeSlash & 384&
        \cellcolor{blue!8}{79.4} & 82.1 & 76.6 & \textbf{95.1} & \textbf{75.9} & \textbf{73.6} & 72.8 & 
        \cellcolor{green!8}{75.2} & 81.4 & 93.7 & 53.9 & \textbf{71.9} \\ 
        
        \cdashline{1-15}
        
        \textbf{\model-0.3B} & Merged-1B~ & Native & \cellcolor{blue!8}{\textbf{80.6}} & 81.5 & \textbf{84.1} & 93.9 & 75.1 & 69.7 & \textbf{79.1} & \cellcolor{green!8}{\textbf{76.3}} & \textbf{84.0} & \textbf{95.1} & \textbf{54.7} & 71.2 \\
        
        \midrule

        OpenCLIP-H~\cite{openclip} &LAION2B-en~ & 224 &
        \cellcolor{blue!8}{72.3} & 78.0 & 59.4 & 89.3 & 70.9 & 66.6 & 69.4 & \cellcolor{green!8}{68.7} & 75.5 & 89.5 & 46.5 & 63.4 \\

        SigLIP-SO400M~\cite{zhai2023sigmoid} & WebLI10B-En~\faEyeSlash & 224 &
        \cellcolor{blue!8}{78.3} & 82.0 & 71.9 & 95.1 & 76.1 & 74.0 & 70.6 & \cellcolor{green!8}{71.9} & 75.3 & 91.0 & 51.8 & 69.7  \\

        MetaCLIP-H~\cite{xu2023demystifying} & CC-2.5B & 224 &
        \cellcolor{blue!8}{78.4} & 80.5 & 75.3 & 93.4 & 74.2 & 70.5 & 76.4 & \cellcolor{green!8}{71.3} & 78.3 & 91.8 & 48.8 & 66.2  \\

        CLIPAv2-H~\cite{li2023clipa} & DataComp1B~ & 336 &
        \cellcolor{blue!8}{80.8} & 81.8 & 82.7 & 94.4 & 75.6 & 72.8 & 77.4 & \cellcolor{green!8}{70.8} & 76.3 & 90.3 & 49.2 & 67.2 \\	
         
        DFN-H~\cite{fang2023data} & DFN5B~\faEyeSlash & 378 &
        \cellcolor{blue!8}{80.5} & \textbf{84.4} & 79.6 & 93.8 & \textbf{78.3}& 73.2 & 73.4 & \cellcolor{green!8}{75.9} & 82.0 & 94.0 & \textbf{55.6} & 71.9 \\

        SigLIP-SO400M~\cite{zhai2023sigmoid} & WebLI10B-En~\faEyeSlash & 384 &
        \cellcolor{blue!8}{81.7} & 83.1 & 82.5 & \textbf{95.8} & 77.2& \textbf{74.5} & 77.0 & \cellcolor{green!8}{76.0} & 83.0 & 94.3 & 54.2 & \textbf{72.4} \\

        \cdashline{1-15}
        
        \textbf{\model-0.6B} & Merged-1B~ & Native & 
        \cellcolor{blue!8}{\textbf{82.1}} & 82.3 & \textbf{86.8} & 94.9 & 76.1 & 71.6 & \textbf{81.1} & \cellcolor{green!8}{\textbf{76.6}} & \textbf{84.1} & \textbf{95.5} & 55.4 & 71.7 \\

        \midrule

        OpenCLIP-g~\cite{openclip} & LAION2B-en~ & 224 &
        \cellcolor{blue!8}{73.0} & 78.5 & 60.9 & 90.2 & 71.6 & 67.5 & 69.1 &
        \cellcolor{green!8}{71.1} & 77.7 & 91.4 & 48.8 & 66.4 \\

        OpenCLIP-G~\cite{openclip} & LAION2B-en~ & 224 &
        \cellcolor{blue!8}{76.2} & 80.1 & 69.3 & 92.1 & 73.6 & 68.9 & 72.8 & \cellcolor{green!8}{72.8} & 79.6 & 92.9 & 51.4 & 67.4 \\

        EVA01-g~\cite{sun2023eva} & Merged-2B~ & 224 &
        \cellcolor{blue!8}{76.9} & 79.3 & 74.2 & 92.5 & 72.1 & 68.1 & 74.9 & \cellcolor{green!8}{72.3} & 79.0 & 91.7 & 50.3 & 68.2 \\

        EVA02-E~\cite{sun2023evaclip} &Merged-2B~ & 336 &
        \cellcolor{blue!8}{80.9} & 82.0 & 82.2 & 94.6 & 75.6 & 71.6 & 79.4 & \cellcolor{green!8}{73.2} & 78.9 & 94.1 & 51.1 & 68.7  \\ 

        CLIPAv2-G~\cite{li2023clipa} & DataComp1B~ & 336 &
        \cellcolor{blue!8}{82.7} & 83.1 & 86.0 & 95.4 & 77.3 & \textbf{74.5} & 79.7 & \cellcolor{green!8}{72.2} & 78.3 & 92.2 & 50.4 & 67.8 \\

        InternViT-6B~\cite{chen2024internvl} & InternVL-5B~ & 224 &
        \cellcolor{blue!8}{82.5} & 83.2 & 83.8 & \textbf{95.7} & 77.3 & 74.3 & 80.6 & \cellcolor{green!8}{75.3} & 81.7 & 94.7 & 54.1 & 70.6 \\

        EVA-8B~\cite{sun2023eva} & Merged-2B~ & 224 &
        \cellcolor{blue!8}{82.9} & \textbf{83.5} & 85.2 & 95.3 & \textbf{77.7} & 74.3 & 81.2 & \cellcolor{green!8}{74.9} & 80.8 & \textbf{95.6} & 53.0 & 70.3 \\

        \cdashline{1-15}
        \textbf{\model-1B} & Merged-1B & Native & \cellcolor{blue!8}{\textbf{83.5}} & 82.9  & \textbf{89.1} & \textbf{95.7} & 77.3 & 73.4 & \textbf{82.8} & \cellcolor{green!8}{\textbf{76.3}} & \textbf{83.5} & 95.1 & \textbf{55.3} & \textbf{71.3} \\
        
        \bottomrule
    \end{tabular}
}
\label{tab:image-zero-shot-eval}
\vspace{-2mm}
\end{table}

\subsubsection{Results Comparison and Analysis}
Table~\ref{tab:image-zero-shot-eval} demonstrates the exceptional performance of our model at comparable parameter scales. As the model size increases from 0.3B to 1.4B, the average zero-shot classification accuracy across six benchmarks exhibits a progressive improvement trend, rising from 80.5\% to 81.9\% and further to 83.4\%. Notably, all models of varying scales employ identical training samples and strategies, with this performance enhancement attributed to \textbf{parameter scaling} effects—a finding consistent with established scaling laws in transformer-related research. As detailed in the table, our \model-1B shows superior performance despite utilizing a smaller training corpus, outperforming its counterparts with more parameters, such as InternViT-6B~\cite{chen2024internvl} and EVA-8B~\cite{sun2023evaclip}. We posit that this advantage stems from two key factors: (1) optimized model atchitecture and training strategy, and (2) preservation of native input resolution, which better preserves the original image aspect ratio to generate higher-quality visual tokens. Furthermore, our model exhibits the most pronounced performance gains on the ImageNet-A benchmark, outperforming the second-ranked model by 4$\sim$7 points, which suggests enhanced robustness against adversarial samples and hallucination artifacts in downstream applications.

\subsection{Results on Zero-shot Video Classification \& Retrieval }
\subsubsection{Evaluation Setup}

We evaluate the zero-shot video classification performance on three popular benchmarks include K-400~\cite{kay2017kinetics}, UCF-101~\cite{soomro2012ucf101} and HMDB51~\cite{kuehne2011hmdb}, using the class names as text prompts. We evaluate the model performance using \textit{Top-1} accuracy.
Also, we evaluate the zero-shot video-text retrieval performance on ActivityNet~\cite{caba2015activitynet}, MSR-VTT~\cite{xu2016msrvtt} and MSVD~\cite{chen2011collecting}. Following~\cite{wang2022internvideo,wang2024internvideo2}, for each video in the 1K version of the test split, we sample one sentence from every set of 20 sentences for MSR-VTT. Following~\cite{rizve2024vidla}, we concatenate the multiple descriptions to form a paragraph and perform a paragraph-to-video retrieval on ActivityNet.
All videos are sampled with a dynamic frame rate, with each frame dynamically resized to maintain the original aspect ratio while ensuring the total token count remains within the range of 576 to 16,384 for model input.

\begin{table}[htb]
\centering
\setlength{\tabcolsep}{2pt}
\renewcommand{\arraystretch}{1.2}
\scriptsize
\caption{
    \textbf{Evaluation of zero-shot performance on various video benchmarks}. The symbol \faEyeSlash~indicates that the video-caption data used by the corresponding method is not publicly available. The blue shading corresponds to the average performance of zero-shot classification tasks, while the green is the average performance of zero-shot retrieval tasks. The $\dagger$ signifies that the reported metrics are based on our own evaluations.
}
\scalebox{0.92}{
    \begin{tabular}{rcccccccccccccc}
        \toprule
        \multirow{2}{*}{\textbf{Method}} & \multirow{2}{*}{\textbf{Type}} & \multirow{2}{*}{\textbf{Res.}} & \multirow{2}{*}{\textbf{Frames}} & \multirow{2}{*}{\textbf{Overall}} & \multicolumn{3}{c}{\textbf{Classification}} & \multirow{2}{*}{\textbf{Overall}}& \multicolumn{2}{c}{\textbf{ANet}} & \multicolumn{2}{c}{\textbf{MSR-VTT}} & \multicolumn{2}{c}{\textbf{MSVD}} \\
        \cdashline{6-8} \cdashline{10-15}
        & & & & & \textbf{K400} & \textbf{UCF} & \textbf{HMDB} & & \textbf{V$\to$T} & \textbf{T$\to$V} & \textbf{V$\to$T} & \textbf{T$\to$V} & \textbf{V$\to$T} & \textbf{T$\to$V} \\
        
        \midrule

        $^{\dagger}$OpenCLIP-L~\cite{openclip} & Image & 224 & 16 &
        \cellcolor{blue!8}{58.4} & 61.5 & 69.2 & 44.5 &
        \cellcolor{green!8}{41.0} & 32.0 & 34.2 & 30.1 & 37.5 & 63.7 & 48.5 \\

        $^{\dagger}$DFN-L~\cite{fang2023data} & Image & 224 & 16 &
        \cellcolor{blue!8}{56.4} & 56.8 & 67.7 & 44.8 &
        \cellcolor{green!8}{40.4} & 31.6 & 34.1 & 32.1 & 35.2 & 61.9 & 47.7 \\

        $^{\dagger}$EVA02-L~\cite{sun2023evaclip} & Image & 336 & 16 &
        \cellcolor{blue!8}{64.4} & 64.4 & 76.0 & 52.8 &
        \cellcolor{green!8}{44.7} & 35.8 & 37.2 & 35.4 & 39.7 & 69.1 & 51.0 \\

        $^{\dagger}$SigLIP-L~\cite{zhai2023sigmoid} & Image & 384 & 16 &
        \cellcolor{blue!8}{64.8} & 64.2 & 79.2 & 50.9 &
        \cellcolor{green!8}{45.3} & 34.3 & 35.8 & 35.7 & 40.0 & 73.0 & \textbf{53.0} \\

        ViCLIP-L~\cite{wang2023internvid} & Video & 224 & 8 &
        \cellcolor{blue!8}{-} & 64.8 & - & - &
        \cellcolor{green!8}{41.2} & 24.0 & 15.1 & 41.3 & 42.4 & 75.1 & 49.1 \\

        InterVideo-L~\cite{wang2022internvideo} & Video & 224 & 16 &
        \cellcolor{blue!8}{-} & 64.3 & 80.5 & - &
        \cellcolor{green!8}{42.2} & 31.4 & 30.7 & 39.6 & 40.7 & 67.5 & 43.4 \\ 

        UMT-L~\cite{liu2022umt} & Video & 224 & 16 &
        \cellcolor{blue!8}{-} & - & - & - &
        \cellcolor{green!8}{47.7} & 39.4 & 41.9 & 38.6 & 42.6 & 74.5 & 49.0 \\
        
        \cdashline{1-15}
        
        \textbf{\model-0.3B} & Image\&Video & Native & 2$\sim$32 &
        \cellcolor{blue!8}{\textbf{68.0}} & \textbf{66.0} & \textbf{82.6} & \textbf{55.4} &  
        \cellcolor{green!8}{\textbf{53.9}} & \textbf{47.9} & \textbf{49.9} & \textbf{48.0} & \textbf{48.8} & \textbf{77.8} & {50.7}\\
        
        \midrule
        
        $^{\dagger}$OpenCLIP-H~\cite{openclip} & Image & 224 & 16 &
        \cellcolor{blue!8}{62.0} & 61.7 & 72.5 & 51.6 &
        \cellcolor{green!8}{43.5} & 36.1 & 38.9 & 34.5 & 38.9 & 63.3 & 49.4 \\

        $^{\dagger}$DFN-H~\cite{fang2023data} & Image & 378 & 16 &
        \cellcolor{blue!8}{62.9} & 63.8 & 76.7 & 48.2 &
        \cellcolor{green!8}{46.2} & 39.7 & 42.9 & 36.1 & 39.6 & 66.6 & 52.4 \\

        $^{\dagger}$SigLIP-SO400M~\cite{zhai2023sigmoid} & Image & 384 & 16 &
        \cellcolor{blue!8}{67.3} & 66.8 & \textbf{83.0} & 52.1 &
        \cellcolor{green!8}{47.5} & 36.6 & 39.3 & 37.5 & 41.1 & 75.5 & \textbf{54.7} \\

        TVTSV2-H~\cite{zeng2023tvtsv2} & Video & 224 & 12 &
        \cellcolor{blue!8}{63.2} & 59.6 & 78.0 & 52.1 &
        \cellcolor{green!8}{-} & - & - & - & 41.3 & - & - \\
        
        \cdashline{1-15}
        
        \textbf{\model-0.6B} & Image\&Video & Native & 2$\sim$32 &
        \cellcolor{blue!8}{\textbf{68.6}} & \textbf{67.6} & 82.9 & \textbf{55.2} &  
        \cellcolor{green!8}{\textbf{54.9}} & \textbf{48.7} & \textbf{51.5} & \textbf{48.6} & \textbf{50.2} & \textbf{75.8} & {54.3} \\

        \midrule

        $^{\dagger}$OpenCLIP-g~\cite{openclip} & Image & 224 & 16 &
        \cellcolor{blue!8}{63.1} & 61.5 & 76.6 & 51.1 &
        \cellcolor{green!8}{44.4} & 36.8 & 39.8 & 36.4 & 39.2 & 64.3 & 50.1 \\

        $^{\dagger}$OpenCLIP-G~\cite{openclip} & Image & 224 & 16 &
        \cellcolor{blue!8}{64.2} & 63.2 & 76.2 & 53.4 &
        \cellcolor{green!8}{46.0} & 36.7 & 41.4 & 36.9 & 41.8 & 67.5 & 51.5 \\

        $^{\dagger}$EVA01-g~\cite{sun2023eva} & Image & 224 & 16 &
        \cellcolor{blue!8}{62.8} & 63.4 & 72.1 & 52.9 &
        \cellcolor{green!8}{45.5} & 37.0 & 40.1 & 37.2 & 40.1 & 67.6 & 50.8 \\

        InternViT-6B~\cite{chen2024internvl} & Image & 224 & 8 &
        \cellcolor{blue!8}{-} & \textbf{69.1} & - & - &
        \cellcolor{green!8}{-} & - & - & 42.4 & 46.3 & - & - \\     
        
        \cdashline{1-15}
        
        \textbf{\model-1B} & Image\&Video & Native & 2$\sim$32 &
        \cellcolor{blue!8}{\textbf{69.0}} & 68.6 & \textbf{81.0} & \textbf{57.3} &  
        \cellcolor{green!8}{54.0} & 47.8 & 49.6 & 48.3 & 47.6 & 75.5 & 55.2 \\

        \midrule

        \fontgray{VideoCoCa-g}~\cite{yan2022videococa} & \fontgray{Video} & \fontgray{224} & \fontgray{8} &
        \fontgray{72.4} & \fontgray{72.0} & \fontgray{86.6} & \fontgray{58.7} &
        \fontgray{39.0} & \fontgray{33.0} & \fontgray{34.5} & \fontgray{64.7} & \fontgray{34.4} & \fontgray{33.0} & \fontgray{34.5} \\

        \fontgray{VideoPrism-g}~\cite{zhao2024videoprism} & \fontgray{Video} & \fontgray{288} & \fontgray{16} &
        \fontgray{-} & \fontgray{76.4} & \fontgray{-} & \fontgray{-} &
        \fontgray{-} & \fontgray{50.3} & \fontgray{52.7} & \fontgray{51.7} & \fontgray{52.7} & \fontgray{-} & \fontgray{-} \\

        \fontgray{InternVideo2-6B}~\cite{wang2024internvideo2} & \fontgray{Video} & \fontgray{224} & \fontgray{8} &
        \fontgray{-} & \fontgray{-} & \fontgray{-} & \fontgray{-} &
        \fontgray{62.0} & \fontgray{56.5} & \fontgray{63.2} & \fontgray{53.7} & \fontgray{55.9} & \fontgray{83.1} & \fontgray{59.3} \\
        
        \bottomrule
    \end{tabular}
}
\label{tab:video-zero-shot-eval}
\vspace{-2mm}
\end{table}

\subsubsection{Results Comparison and Analysis}
Table~\ref{tab:video-zero-shot-eval} shows the performance of our \model~series models on video benchmarks across comparable parameter scales. As the model size scales from 0.3B to 1B, \model~exhibits consistent performance gains on video benchmarks, with average zero-shot classification metrics improving from 68.0 to 69.0. When compared to models trained on image-caption data under similar parameter scales, \model~achieves notable improvements. These advancements can be attributed to two key design choices: (1) preserving the aspect ratio of each frame to retain the original semantic information of visual content, and (2) employing dynamic video frame sampling to effectively capture detailed temporal information. However, when compared to the models trained exclusively on video-caption data, \model~still has room for improvement compared to some of the latest models~\cite{yan2022videococa,zhao2024videoprism,wang2024internvideo2}, as shown in the Table~\ref{tab:video-zero-shot-eval} with gray color. This suggests that while integrating image data helps maintain image-related capabilities, it may partially compromise the model's ability to fully leverage temporal information in videos.

\subsection{Results on Image Classification by Linear Probing}

Following common prectices \cite{chen2024internvl, fini2024multimodal}, we assess the performance of \model~family as off-the-shelf backbones on image classifications. Specifically, we train a linear classifier on the last feature layer with a frozen backbone on ImageNet-1K \cite{deng2009imagenet} and evaluate the performance on the validation set and other ImageNet variants ~\cite{beyer2020imagenetreal,hendrycks2021natural,hendrycks2021many,recht2019imagenetv2,wang2019learning}. In addition, we also report the classification performance with  attentive probing setting as used in \cite{fini2024multimodal}, which adopts a cross-attention layer with random initialized queries. 
Table \ref{tab:image-eval-classification} represents the downstream classification performance of our models.
First, as the model size increases, the average performance across six benchmarks demonstrates consistent improvement.
Second, we observe that the attentive probing performance shows stable improvements over linear probing. This is reasonable as the attentive head possesses more learnable parameters to adapt to downstream tasks. 
Furthermore, compared to public methods, our \model~shows superior performance across various parameter scales. It is notable that our \model-0.6B outperforms InternViT-6B \cite{chen2024internvl} over 2 points on average performance under the same setting, which demonstrates the robustness and high-quality of visual representation produced by our \model~family.

\begin{table}
	\centering
	\setlength{\tabcolsep}{8pt}
	\renewcommand{\arraystretch}{1.2}
	\scriptsize
	\caption{
		\textbf{Evaluation of classification performance on various image benchmarks}. 
	}
	\scalebox{0.9}{
		\begin{tabular}{rlcccccccc}
			\toprule
            \multirow{2}{*}{\textbf{Method}} & \multirow{2}{*}{\textbf{Classifier}} & \multirow{2}{*}{\textbf{Res.}} & \multirow{2}{*}{\textbf{Overall}} & \multicolumn{6}{c}{\textbf{ImageNet Variants}} \\
            \cdashline{5-10}
			& & & & \textbf{IN-1K} & \textbf{IN-Real} & \textbf{IN-V2} & \textbf{IN-A} & \textbf{IN-R} & \textbf{IN-S} \\

		\midrule
            NaViT-L~\cite{dehghani2023patch} & Linear & 224 &
			{-} & 76.0 & {-} & {-} & 65.5 & {-} & {-} \\
            CLIP-L~\cite{radford2021clip} & Linear & 336 &
			{-} & 85.3 & 88.8 & 75.8 & {-} & {-} & {-} \\

            SigLIP-L~\cite{zhai2023sigmoid} & Attentive & 224 &
			{-} & 86.5 & - & - & {-} & {-} & {-} \\

            AIMv2-ViT-L~\cite{fini2024multimodal} & Attentive & 224 &
			{-} & 86.6 & {-} & {-} & {-} & {-} & {-} \\
		    
			\cdashline{1-10}
		\textbf{\model-0.3B} & Linear & Native & 
			{83.0} & 87.6 & {90.3} & {79.5} & \textbf{84.1} & 90.6 & {66.0} \\
            
            \textbf{\model-0.3B} & Attentive & Native & 
			\textbf{83.3} & \textbf{87.7} & \textbf{90.5} & \textbf{79.8} & 83.8 & \textbf{91.1} & \textbf{66.8} \\
			\midrule
			 
            CLIP-H~\cite{radford2021clip} & Linear & 224 &
		{-} & 84.4 & 88.4 & 75.5 & {-} & {-} & {-} \\
            
            $^{\dagger}$DFN-H~\cite{fang2023data} & Linear & 378 &
		{81.6} & 87.3 & 90.4 & 78.8 & {74.8} & {90.3} & {68.3} \\

            SigLIP-SO400M~\cite{zhai2023sigmoid} & Attentive & 384 &
		{-} & 87.3 & - & - & {-} & {-} & {-} \\

            AIMv2-ViT-H~\cite{fini2024multimodal} & Attentive & 224 &
		{-} & 87.5 & {-} & {-} & {-} & {-} & {-} \\
			\cdashline{1-10}
			
		\textbf{\model-0.6B} & Linear & Native & 
			{84.4} & 88.2 & {90.6} & {80.6} & 87.1 & 92.0 & {68.0} \\
            
            \textbf{\model-0.6B} & Attentive & Native & 
			{\textbf{84.8}} & \textbf{88.3} & \textbf{90.7} & \textbf{81.0} & \textbf{87.3} & \textbf{92.5} & \textbf{68.8} \\
	
		\midrule

            OpenCLIP-G~\cite{openclip} & Linear & 224 &
             78.5 & 86.2 & 89.4 & 77.2 & 63.8 & 87.8 & 66.4 \\
            
            DINOv2-g~\cite{oquab2023dinov2} & Linear & 224 &
            78.6 & 86.5 & 89.6 & 78.4 & 75.9 & 78.8 & 62.5 \\
            
            EVA01-g~\cite{sun2023eva} & Linear & 224 &
            79.1 & 86.5 & 89.3 & 77.4 & 70.5 & 87.7 & 63.1 \\

            AIMv2-ViT-1B~\cite{fini2024multimodal} & Attentive & 224 &
			{-} & 88.1 & {-} & {-} & {-} & {-} & {-} \\
 
            InternViT-6B~\cite{chen2024internvl} & Linear & 224 &
            {82.5} & 88.2 & 90.4 & 79.9 & 77.5 & 89.8 & 69.1 \\

            EVA-8B~\cite{sun2023eva} & Linear & 224 &
            {-} & 88.5 & {-} & {-} & {-} & {-} & {-} \\

            \cdashline{1-10}

            \textbf{\model-1B} & Linear & Native & \textbf{86.0} & 88.9 & 90.8 & 81.5 & 90.1 & \textbf{94.0} & \textbf{70.7} \\
            \textbf{\model-1B} & Attentive & Native & \textbf{86.0} & \textbf{{89.2}} & \textbf{91.0} & \textbf{81.7} & \textbf{90.1} & 93.6 &
            70.6\\
            
			\bottomrule
		\end{tabular}
}
	\label{tab:image-eval-classification}
	\vspace{-2mm}
\end{table}

\subsection{Results on Dense prediction.}
In this section, we evaluate the dense prediction performance of our \model~family by transferring to semantic segmentation. Following \cite{chen2024internvl, dehghani2023vit22b}, we fine-tune a decoder with freezing backbones under two different structures, \textit{i.e.}, Linear and UperNet. Linear decoder transforms the dimension of one single layer visual feature to number of semantic classes, while the UperNet decoder employs PPM and FPN to integrates multi-scale features. Experiments are conducted on the ADE20K \cite{zhou2017ade20k} dataset. In terms of data preprocessing, we employed the same fixed-resolution input and data augmentation strategies as those used in InternViT \cite{chen2024internvl}.
Corresponding results are shown in Table \ref{tab:seg-performence}. We can observe a performance gap between these two types of decoder, this can be understand that UperNet has significantly more trainable parameters than Linear decoder. Taking \model-0.6B as an example, Linear decoder has a  parameter count of 0.2M, whereas UperNet contains approximately 200M parameters. Notably, our \model~Family demonstrates an obvious performance advantage compared with existing state-of-the-art vision encoders. Under the setting of Linear decoder, our \model-1B achieves a performance of $45.4$ mIoU, which is $+6.1$ points over OpenCLIP-G \cite{openclip} and $+10.8$ points over ViT-22B \cite{dehghani2023scaling}. In the case of UperNet decoder, our \model-1B reaches 56.2 mIoU, also surpassing larger parameter-scale model like InternViT-6B \cite{chen2024internvl}. 

\begin{table}[htb]
    \centering
    \setlength{\tabcolsep}{0.5cm}
    \caption{\textbf{Evaluation of semantic segmentation on ADE20k dataset with frozen backbones.}
    }
    \scalebox{0.725}{
        \begin{tabular}{rcccc}
        \toprule
        \textbf{Method} & \textbf{CropSize} & \textbf{mIoU$^{Linear}$} & \textbf{mIoU$^{UperNet}$} \\
        \midrule
        CLIP-L \cite{radford2021clip} & {-} & 39.0 & {-} \\
        SigLIP-SO400M~\cite{zhai2023sigmoid} & {-} & 40.8 & {-} \\
        $^{\dagger}$DFN-H~\cite{fang2023data} & {-} & 41.3 & {-}\\
        OpenCLIP-G \cite{openclip} & ${512^{2}}$ & 39.3 \\
        InternViT-6B \cite{chen2024internvl} &  ${504^{2}}$ & \textbf{47.2} & 54.9\\
        ViT-22B \cite{dehghani2023scaling}  & ${504^{2}}$ & 34.6 & 52.7 \\
        \cdashline{1-5}
        \textbf{\model-0.3B} & ${504^{2}}$   &  40.7 & 54.6\\
        \textbf{\model-0.6B} & ${504^{2}}$   &   42.9 & 55.1 \\
        \textbf{\model-1B} & ${504^{2}}$   &   45.4 & \textbf{56.2}\\
        \bottomrule
        \end{tabular}
    }
    \label{tab:seg-performence}
\end{table}

\subsection{Results on Multimodal Understarding}
\subsubsection{Evaluation Setup}
To assess the potential of multimodal understanding, we employ a dual-stage training paradigm, similar to common practices~\cite{chen2024allava,chen2023sharegpt4v}. 
In the pretraining stage, we train the projector with a learning rate of $1e^{-3}$ using a merged $2.3$M dataset comprised of LLaVA-CC3M-Pretrain~\cite{liu2023llava}, ALLaVA-Caption~\cite{chen2024allava}, ShareGPT4V-PT~\cite{chen2023sharegpt4v}. 
In the fine-tuning stage, we unfreeze the whole model, and train it with a learning rate of $1e^{-5}$, using the high-quality instrution-tuning dataset LLaVA1.5-Finetune~\cite{liu2023improved}. Note that the native-resolution strategy of \textit{Boundary Markers} and \textit{Line Anchors} are only applied in the fine-tuning stage. 
We employ Qwen2.5-1.5B~\cite{bai2025qwen25vltechnicalreport} as the language branch of the multimodal language model due to its excellent efficiency.

All evaluations are conducted using VLMEvalKit~\cite{duan2024vlmevalkit}, assessing performance across 16 popular benchmarks, including
GQA~\cite{hudson2018gqa}, 
DocVQA~\cite{mathew2021docvqa}, 
InfoVQA~\cite{mathew2022infographicvqa}, 
ScienceQA~\cite{lu2022learn},
TextVQA~\cite{singh2019textvqa},
VizWiz~\cite{gurari2018vizwiz},
OCRVQA~\cite{mishraICDAR19},
OCRBench~\cite{liu2024ocrbench},
MME~\cite{fu2023mme},
MMMU~\cite{yue2024mmmu},
SEEDBench\_IMG~\cite{li2023seed},
MathVista\_MINI~\cite{lu2023mathvista},
AI2D~\cite{kembhavi2016diagram},
HallusionBench~\cite{guan2024hallusionbench},
POPE~\cite{li2023evaluating},
HRBench4K~\cite{hrbench}.

\subsubsection{ Results Comparison and Analysis}
As illustrated in Table~\ref{tab:mm-performence}, under exactly the same training data and training strategy, the proposed \model~surpasses various state-of-the-art vision encoders~\cite{fang2023data, fini2024multimodal, zhai2023sigmoid} on numerous multimodal understanding benchmarks.
Notably, ~\model~demonstrates exceptional capabilities in scenarios involving dense information, such as document parsing~\cite{mathew2021docvqa}, graphic parsing~\cite{mathew2022infographicvqa}, and high-resolution tasks~\cite{hrbench}. We argue that the native resolution plays a crucial role in achieving outstanding performance in these areas, ensuring minimal loss of image information.
We also assess the effectiveness of the proposed strategy of \textit{Boundary Markers} and \textit{Line Anchors} on 0.6B model size, as demonstrated in Table~\ref{tab:mm-performence}, which highlights their impact.

\begin{table}[htb]
\centering
\setlength{\tabcolsep}{5pt}
\renewcommand{\arraystretch}{1.2}
\scriptsize
\caption{
    \textbf{Evaluation of multimodal understanding on various vision-language benchmarks}. Note the superscript $\triangle$ represents model with \textit{Boundary Markers} and \textit{Line Anchors}.
}
\scalebox{0.99}{
    \begin{tabular}{lcccccccc}
    \toprule
    \multirow{2}{*}{\textbf{Benchmarks}} & \multirow{2}{*}{\textbf{SigLIP-L}~\cite{zhai2023sigmoid}} & \multirow{2}{*}{\textbf{DFN-H}~\cite{fang2023data}} & \multirow{2}{*}{\textbf{AIMv2-H}~\cite{fini2024multimodal}} & \multirow{2}{*}{\textbf{SigLIP-SO400M}~\cite{zhai2023sigmoid}} & \multicolumn{4}{c}{\textbf{\model}} \\
    \cdashline{6-9}
    & & & & & \textbf{0.3B}{$^\triangle$} & \textbf{0.6B} & \textbf{0.6B}{$^\triangle$} & \textbf{1B}{$^\triangle$} \\
    
    \midrule
    Resolution & 378 & 378 & 448 & 384 & \multicolumn{4}{c}{\textbf{Native}} \\
    \cdashline{1-9}
    
    GQA$_{TestDev\_Balanced}$ & 61.5 & 60.6 & 61.5 & 61.0 & 60.8 & 58.2 & 60.3 & 61.2 \\
    DocVQA$_{VAL}$ & 30.8 & 25.9 & 36.2 & 32.0 & 47.7 & 46.3 & 48.2 & 47.0 \\
    InfoVQA$_{VAL}$ & 22.7 & 22.1 & 25.8 & 23.2 & 27.8 & 28.0 & 28.5 & 27.5 \\
    ScienceQA$_{VAL}$ & 63.6 & 62.7 & 64.8 & 66.4 & 64.5 & 65.0 & 63.6 & 65.3 \\
    TextVQA$_{VAL}$ & 48.0 & 41.7 & 53.2 & 50.9 & 50.8 & 52.0 & 50.7 & 52.0 \\
    VizWiz & 30.5 & 28.5 & 30.3 & 30.8 & 29.1 & 29.8  & 29.4 & 29.3 \\
    OCRVQA$_{TESTCORE}$ & 31.2 & 32.0 & 31.0 & 30.9 & 32.2 & 31.6 & 32.2 & 32.1 \\
    OCRBench & 35.2 & 30.6 & 22.4 & 36.0 & 33.6 & 37.0 & 36.9 & 36.4 \\
    MME & 59.3 & 62.6 & 59.8 & 60.0 & 57.9 & 58.6  & 59.0 & 60.7 \\
    MMMU$_{VAL}$ & 35.9 & 34.2 & 37.1 & 35.4 & 36.1 & 38.7 & 36.6 & 37.0 \\
    SEEDBench$_{IMG}$ & 70.0 & 70.3 & 70.9 & 71.2 & 68.2 & 67.8 & 68.0 & 69.3 \\
    MathVista$_{MINI}$ & 28.6 & 29.9 & 30.1 & 29.6 & 27.5 & 27.9 & 28.5 & 28.7 \\
    AI2D$_{Test}$ & 60.5 & 58.3 & 60.4 & 60.6 & 58.0 & 57.7 & 58.7 & 59.3 \\
    HallusionBench & 56.8 & 56.6 & 53.9 & 54.8 & 57.0 & 55.2 & 57.8 & 54.1 \\
    POPE & 87.2 & 88.0 & 85.4 & 87.7 & 87.1 & 88.1 & 87.9 & 88.1 \\
    HRBench4K & 39.9 & 39.5 & 44.5 & 45.0 & 44.1 & 43.6 & 46.1 & 46.4 \\
    \cdashline{1-9}
    
    Average & 47.6 & 46.5 & 48.0 & 48.5 & 48.9 & 49.1 & 49.5 & 49.6 \\
    
    \bottomrule
    \end{tabular}
}
\label{tab:mm-performence}
\end{table}

\subsection{Ablation Study}
\subsubsection{Robustness verification of resolution mode.}
\begin{figure}[htb]
\centering
\scalebox{0.93}{
    \includegraphics[width=\textwidth]{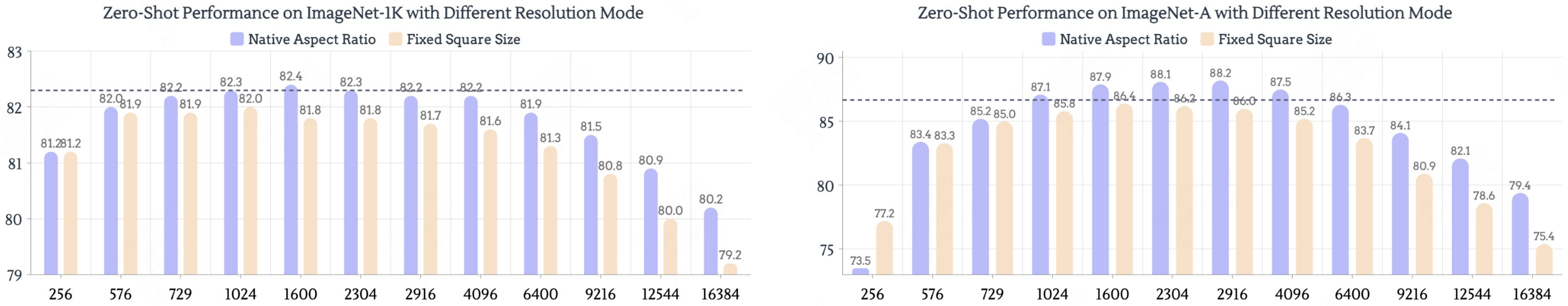}
}
\caption{
    \textbf{Performance comparison of different resolution modes as the length of the vision sequence increases}. The black dashed line shows the performance when using native resolution.} 
\label{fig:univitar_resolution_mode}
\end{figure}

In this section, we analyze the performance of three resolution modes (fixed resolution, native aspect ratio, and native resolution) across varying visual sequence lengths in Figure~\ref{fig:univitar_resolution_mode}. For fixed resolution mode, following common practices, we resize the shorter edge of each image to a predefined size $S$ and apply \textit{CenterCrop} to ensure the sequence length strictly equals $(S/14)^2$, where $14$ represents the model's patch size. Increasing $S$ proportionally extends the sequence length. 
In native aspect ratio mode, we scale images while preserving their original width-height ratios, ensuring that $wh/14^2$ approximates the target sequence length. We evaluate 12 sequence lengths ranging from 256 to 16,384 tokens, testing zero-shot classification performance on ImageNet-1K and ImageNet-A using the \model-0.6B model. The experimental results reveal three key findings: \textit{1)} performance initially improves then declines with increasing sequence lengths under both fixed and native aspect ratio modes, peaking at 1024$\sim$4096 token lengths. \textit{2)} native aspect ratio mode consistently outperforms fixed resolution, indicating that preserving original aspect ratios better retains image information during inference. \textit{3)} native aspect ratio mode occasionally surpasses native resolution performance at certain sequence lengths.

\subsubsection{Verification of the effectiveness of training strategies.}
\begin{figure}[htb]
\centering
\scalebox{0.93}{
    \includegraphics[width=\textwidth]{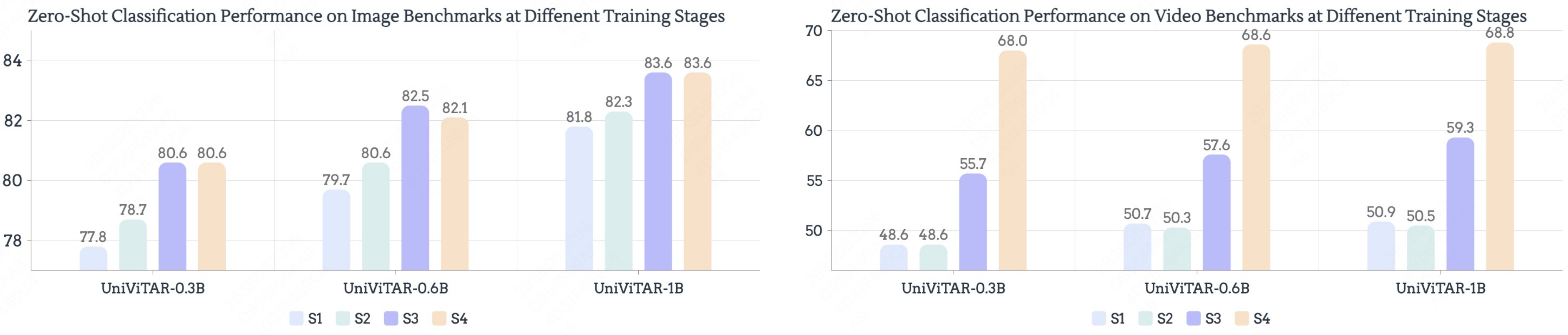}
}
\caption{
    \textbf{Average performance improvement illustration across different training stages}. 
    } 
\label{fig:univitar_train_strategy}
\end{figure}

As introduced in the methodology section, we categorize the training strategies for our \model~into four distinct stages.
As shown in Figure~\ref{fig:univitar_train_strategy}, for zero-shot classification tasks on image benchmarks (left part), we show that S1, S2, and S3 exhibit progressive performance improvements, while phase S4 maintains comparable accuracy despite incorporating image-video alternating training. In contrast, for zero-shot video classification (right part), S1 and S2 show minimal performance variation, with dynamic-resolution training in phase S3 significantly boosting video understanding capabilities, followed by further enhancements in S4 through image-video cross-modal training. This demonstrates that dynamic-resolution training enables models to process more native visual sequences, while the final unified training phase equips the model with generalized capabilities for handling diverse visual modalities. 

\subsubsection{Verification of the effectiveness of image-video alternative strategy.}
To validate the efficacy of the alternating image-video training strategy, we conducted initial experiments with 100M image-text pairs and 10M video-text pairs. Note that the image-to-video data ratio is approximately 10:1, consistent with the ratio used in stage 4 of the \model~series. We trained a \model-0.3B model for 3 epochs, comparing mixed training and alternating training strategies.
As shown in Table~\ref{tab:ablation_mixed_alter_strategy}, the alternating training strategy outperforms the mixed strategy across key image and video benchmark metrics, demonstrating its effectiveness in enhancing visual representation learning. This performance gain can be attributed to the increased training difficulty arising from the unification of data modalities within each batch.

\begin{table}[htbp]
    \centering
    \caption{\textbf{Zero-shot classification performance of image-video training strategy}. 
    }
    \scalebox{0.8}{
        \begin{tabular}{lcccc}
            \midrule
            Strategy & ImageNet-1K & ImageNet-A & K400 & UCF101 \\
            \midrule
            Batch-Mixed & 70.46 & 45.89 & 58.82 & 75.15 \\
            Batch-Alternative & 71.25 & 48.60 & 61.01 & 77.66 \\
            \midrule
        \end{tabular}
    }
    \label{tab:ablation_mixed_alter_strategy}
\end{table}

\subsubsection{Verification of effectiveness of native resolution for video.}
\begin{figure}[htb]
\centering
\resizebox{0.88\textwidth}{!}{\includegraphics{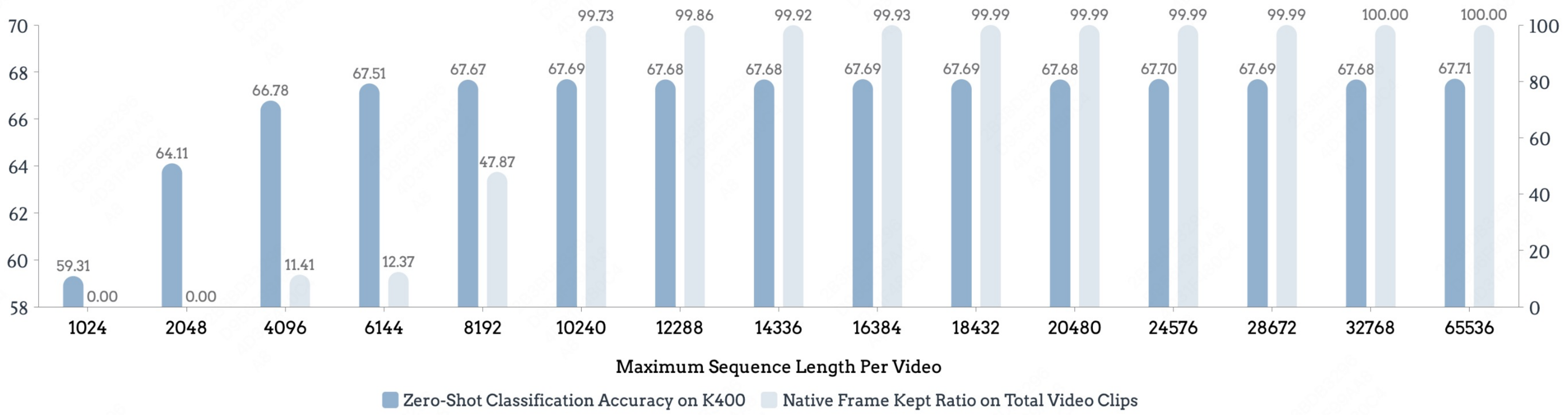}}
\caption{
    \textbf{Performance changes on the K400 dataset across varying sequence length limits.}
    }
\label{fig:univitar_native_resolution}
\end{figure}

In this section, we conduct an ablation study to explore the role of native resolution in video data processing. We dynamically sample a maximum of 32 frames (denoted as $F$) for each video clip. For frames exceeding the sequence length limit, we resize them while preserving their native aspect ratio to a smaller resolution.
We evaluate 15 maximum video sequence length, ranging from 1024 to 65,536, and test the zero-shot classification performance of \model-0.6B on the K400 dataset.
Note that the minimum video sequence length is fixed to 576.
As shown in Figure~\ref{fig:univitar_native_resolution}, the performance initially improves and then stabilizes as the sequence length limit increases, reaching a plateau at length 10,240. We attribute this to the fact that, with 32 sampled frames, a sequence length of 10,240 corresponds to a resolution of $490\times256$, enabling most frames in K400 to retain their native resolution during data processing. This finding underscores the importance of native resolution in enhancing video understanding capabilities.

\subsubsection{Verification of effectiveness of data scale.}

From an intuitive perspective, data scale has a significant impact on the effectiveness of contrastive learning. In this section, we conduct cold-start experiments on \model-0.3B to confirm this view. For the experiment setup, seen samples is fixed at 1B. We respectively train the \model-0.3B for 1 epoch using Merged-1B and for 10 epochs using Merged-100M, which contains 100M image-text pairs that randomly sampled from Merged-1B. Result on zero-shot classification and retrieval is shown in Table~\ref{tab:data-scale-abl}. There is an observable trend where the performance improves as the dataset scale increases. With larger dataset scale, the model is exposed to a broader range of image-text pairs, facilitating a more comprehensive learning and understanding of the visual and linguistic space, thereby enhancing zero-shot performance.

\begin{table}[htb]
\centering
\setlength{\tabcolsep}{3pt}
\renewcommand{\arraystretch}{1.2}
\scriptsize
\caption{
    \textbf{Ablation results of \model-0.3B under varying data scale.}
}
\scalebox{0.93}{
    \begin{tabular}{lccccccccccccc}
        \toprule
        \multirow{2}{*}{\textbf{Data}} & \multirow{2}{*}{\textbf{Seen Samples}}  & \multirow{2}{*}{\textbf{Overall}} & \multicolumn{6}{c}{ImageNet Variants} & \multirow{2}{*}{\textbf{Overall}}& \multicolumn{2}{c}{Flickr} & \multicolumn{2}{c}{COCO} \\
        \cdashline{4-9} \cdashline{11-12} \cdashline{13-14}
        & & & \textbf{IN-1K} & \textbf{IN-A} & \textbf{IN-R} & \textbf{IN-V2} & \textbf{IN-S} & \textbf{O-Net} & & \textbf{T$\to$I} & \textbf{I$\to$T} & \textbf{T$\to$I} & \textbf{I$\to$T} \\

        \midrule

        Merged-100M & 1B & 
      {60.8} & 69.7 & 39.6 & 79.0 & 61.6
       &  54.8 & 60.2 & {67.0} & 73.4 & 88.3 & 44.2 &  62.0 \\
       
        Merged-1B & 1B & {64.2} & 71.7 & 45.7 & 82.3 & 64.3 & 57.3 & 63.9 & {68.9} & 74.9 & 90.7 & 46.0 & 63.8 \\

        \midrule
    \end{tabular}
}
\label{tab:data-scale-abl}
\vspace{-2mm}
\end{table}

\section{Conclusion}
In this work, we introduce \model, a family of homogeneous vision foundation models tailored for unified visual modality and native-resolution scenarios in the era of multimodal. By integrating advanced architectural upgrades into the vanilla ViT paradigm along with incorporating a hybrid training framework—combining resolution curriculum learning, visual feature distillation, and inter-batch modality adaptation—\model~achieves significant improvements across diverse tasks, spanning image/video zero-shot classification/retrieval, dense prediction accuracy, and vision-language model transfer performance. Notably, all models are trained exclusively on public-accessible datasets, where we observe consistent performance gains with parameter scaling from 0.3B to 1B. We hope that our \model~offers the community a versatile framework for advancing multimodal research.

\bibliographystyle{abbrvnat}
\bibliography{main}

\end{CJK*}
\end{document}